\documentclass[]{bytedance_seed}

\usepackage{minitoc}
\usepackage{url}
\usepackage{amssymb}
\usepackage[utf8]{inputenc}
\usepackage{microtype}
\usepackage{booktabs}
\usepackage{pifont} 
\usepackage{multirow}
\usepackage{makecell}
\usepackage{paralist}
\usepackage{xspace}
\usepackage{color}
\usepackage{xcolor}
\usepackage{colortbl}
\usepackage{adjustbox}
\usepackage{hyperref} 
\usepackage[edges]{forest}
\usepackage{tikz} 
\usepackage{caption}
\usepackage{amsfonts}
\usepackage[toc,page,header]{appendix}
\usepackage[utf8]{inputenc}
\usepackage{bbm}
\usepackage{enumitem}

\definecolor{seedc}{RGB}{7, 92, 173}

\newcommand{\name}[1]{GR-3}

\newcommand{\hardware}[1]{ByteMini}
\newcommand{\bo}{\mathbf{o}}
\newcommand{\ba}{\mathbf{a}}
\newcommand{\bs}{\mathbf{s}}

\renewcommand{\paragraph}[1]{\vspace{0.1em}\noindent\textbf{#1}}

\title{GR-3 Technical Report}

\author[]{ByteDance Seed}

\contribution[]{Full Author List in \hyperref[sec:contributions]{Contributions and Acknowledgements}}

\abstract{
We report our recent progress towards building generalist robot policies, the development of \name{}.
\name{} is a large-scale vision-language-action (VLA) model.
It showcases exceptional capabilities in generalizing to novel objects, environments, and instructions involving abstract concepts.
Furthermore, it can be efficiently fine-tuned with minimal human trajectory data, enabling rapid and cost-effective adaptation to new settings.
\name{} also excels in handling long-horizon and dexterous tasks, including those requiring bi-manual manipulation and mobile movement, showcasing robust and reliable performance.
These capabilities are achieved through a multi-faceted training recipe that includes co-training with web-scale vision-language data, efficient fine-tuning from human trajectory data collected via VR devices, and effective imitation learning with robot trajectory data.
In addition, we introduce ByteMini, a versatile bi-manual mobile robot designed with exceptional flexibility and reliability, capable of accomplishing a wide range of tasks when integrated with \name{}.
Through extensive real-world experiments, we show \name{} surpasses the state-of-the-art baseline method, $\pi_0$, on a wide variety of challenging tasks.
We hope \name{} can serve as a step towards building generalist robots capable of assisting humans in daily life.
}

\date{\today}
\correspondence{\email{wuhongtao.123@bytedance.com}}

\checkdata[Project Page]{\url{https://seed.bytedance.com/GR3}}

\begin{document}
\maketitle

\section{Introduction}
\label{sect:intro}

The pursuit of intelligent generalist robots that are capable of assisting humans with daily tasks has been a long-standing goal in robotics research~\cite{brohan2022rt, brohan2023rt, black2410pi0, team2025gemini, bjorck2025gr00t, cheang2024gr, barreiros2025careful}.
A key challenge stems from the immense diversity of the real world, requiring robot policies to possess strong generalization capabilities to handle a wide range of novel scenarios.
Additionally, many daily tasks are inherently long-horizon and require complex dexterous manipulation, demanding robot policies to be highly robust and reliable.

Recent advances in Vision-Language-Action (VLA) models~\cite{brohan2023rt, black2410pi0, intelligence2025pi_, team2025gemini, kimopenvla} have paved a promising path towards developing intelligent generalist robot policies.
These models are built upon pre-trained vision-language models (VLMs)~\cite{beyer2024paligemma,chen2023pali,driess2023palm,li2024llava} and integrate the action prediction capabilities, enabling robots to perform a wide range of tasks following natural language instructions.
Despite these progresses, instruction following remains a significant challenge, especially for out-of-distribution instructions that involve novel object categories that are unseen in robot trajectory data and/or complex concepts that require sophisticated reasoning~\cite{liu2024rdt,brohan2023rt}.
Moreover, VLA models typically require large amounts of demonstrations for policy training, posing substantial challenges for efficient adaptation to novel settings.
Finally, ensuring robustness in complex long-horizon tasks remains challenging due to accumulative errors, especially for tasks involving dexterous skills, such as manipulating deformable objects.

\begin{figure}[!t]
    \centering
    \vspace{-0.5cm}
    \includegraphics[width=0.86\linewidth]{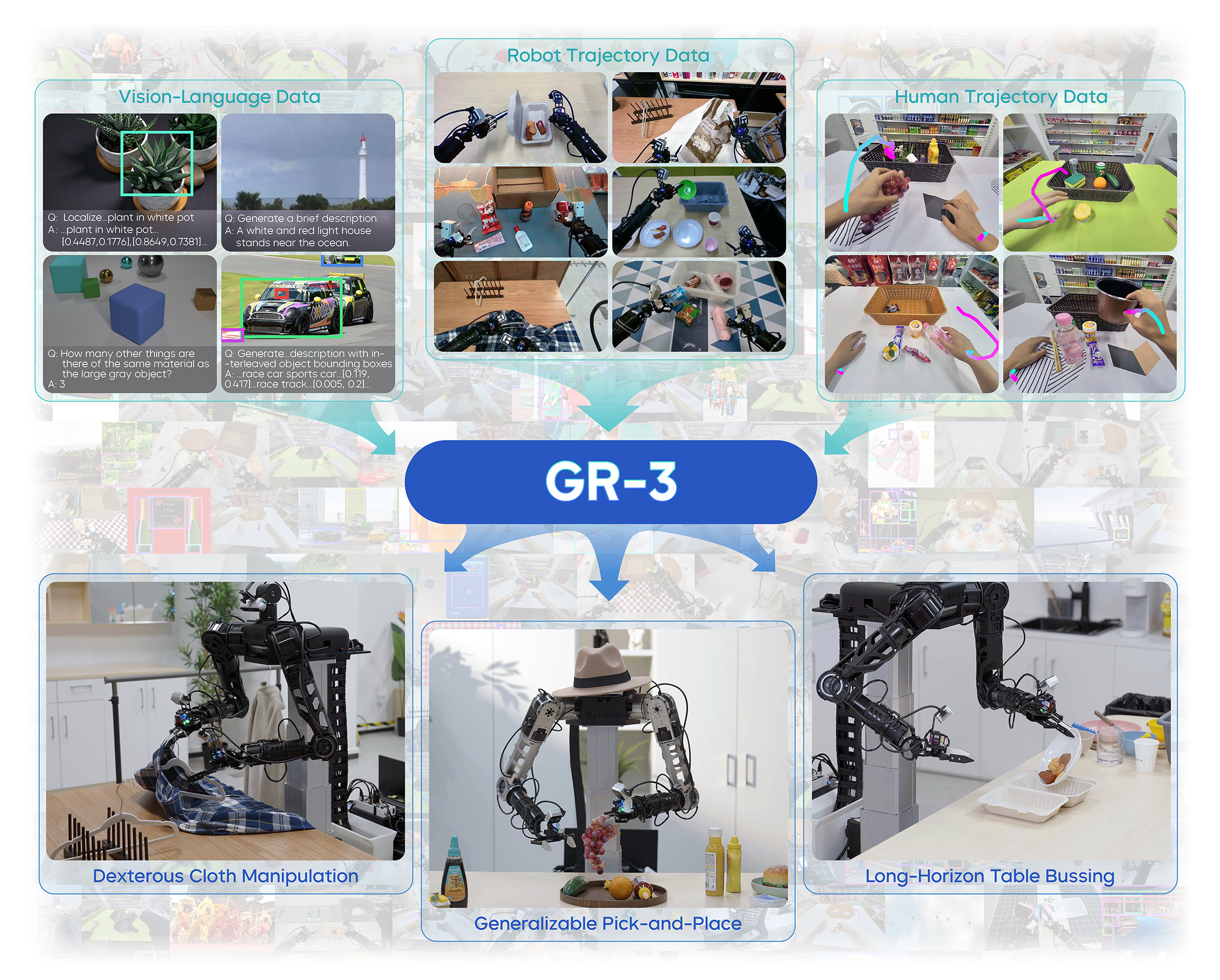}
    \vspace{-0.1cm}
    \caption{
        \textbf{Overview.}
        \name{} is able to learn from three types of data: vision-language data, robot trajectory data, and human trajectory data.
        It is able to perform dexterous and long-horizon tasks with exceptional robustness and generalize well to novel objects, environments, and instructions.
    }
    \vspace{-0.3cm}
    \label{fig:teaser}
\end{figure}

\begin{figure}
    \vspace{-1cm}
    \centering
    \includegraphics[width=0.9\linewidth]{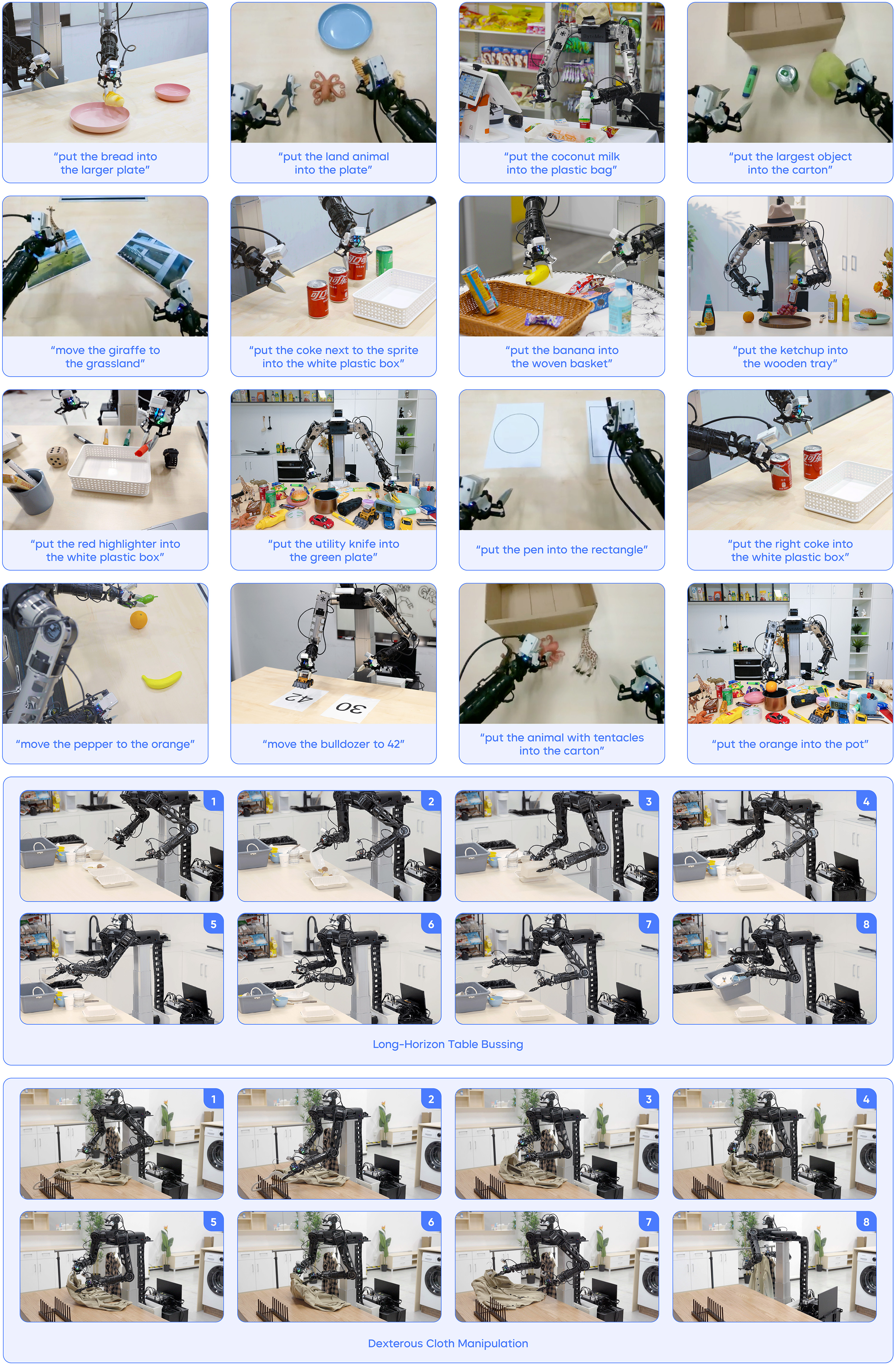}
    \caption{
        \textbf{Capabilities.}
        \name{} strictly follows instructions and is capable of understanding unseen instructions involving abstract concepts.
        It performs robustly and reliably on long-horizon table bussing and dexterous cloth manipulation.
    }
    \label{fig:capability}
\end{figure}

In this report, we introduce \name{}, a large-scale vision-language-action (VLA) model that 1) strictly follows languages and generalizes well to novel objects, environments, and instructions, 2) efficiently learns from few-shot human trajectory data for rapid adaptation to novel settings, and 3) performs long-horizon and dexterous tasks with high robustness and reliability (Fig.~\ref{fig:teaser}).
\name{} takes as inputs a natural language instruction, observations from the environment, and the robot state.
It outputs an action chunk to control a bi-manual mobile robot in an end-to-end manner.
Specifically, \name{} is built upon a pre-trained VLM~\cite{bai2025qwen2} and predicts actions via flow-matching~\cite{lipman2022flow, liu2022rectified}.
We perform careful studies on model architecture, and introduce a set of well-chosen design choices, which we find crucial for instruction following capabilities and long-horizon task performance.
To enhance generalization capabilities, we co-train \name{} on robot trajectory data and large-scale vision-language data covering a wide variety of vision-language tasks.
This training recipe enables \name{} to not only handle objects from novel categories, but also understand abstract concepts relating to sizes, spatial relationships, and common-sense knowledge (Fig.~\ref{fig:capability}), which are unseen within the robot trajectory data.
Besides, we showcase that \name{} can be efficiently fine-tuned with minimal human trajectory data collected via VR devices, enabling rapid and cost-effective adaptation to novel settings.
Alongside \name{}, we introduce ByteMini, a versatile bi-manual mobile robot designed with high flexibility and reliability, capable of accomplishing a wide range of challenging tasks in the real world.

We perform extensive real-world experiments across three challenging tasks: 1) generalizable pick-and-place, 2) long-horizon table bussing, and 3) dexterous cloth manipulation.
\name{} consistently outperforms the state-of-the-art baseline, $\pi_0$, across all tasks.
It showcases strong capabilities in generalizing to objects from novel categories and understanding complex semantics.
Moreover, it efficiently adapts to novel objects with only 10 human trajectories per objects.
Finally, \name{} excels in performing long-horizon and dexterous tasks with remarkable robustness, achieving high average task progress in the challenging tasks of table bussing and cloth manipulation.

\begin{figure}[!t]
    \centering
    \includegraphics[width=\linewidth]{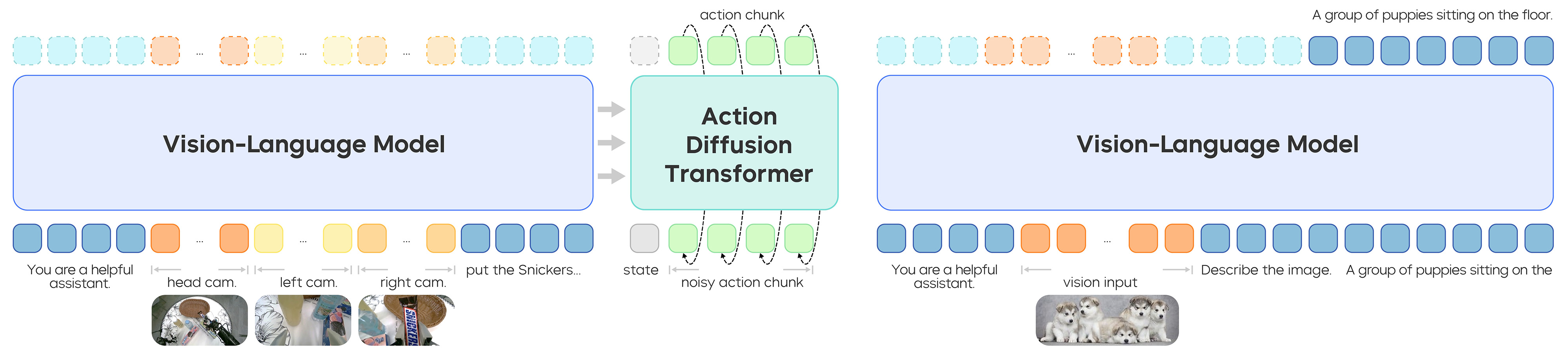}
    \caption{\textbf{The \name{} Model.} \name{} is co-trained on both robot trajectories and vision-language data with a flow-matching objective (left) and a next-token-prediction objective (right), respectively.}
    \label{fig:arch}
\end{figure}

\section{The \name{} Model}
\label{sec:gr3_model}
\name{} is an end-to-end vision-language-action (VLA) model $\pi_{\theta}$.
It controls a bi-manual robot with a mobile base by generating a $k$-length action chunk $\ba_t  = a_{t:t+k}$ conditioned on the input language instruction $l$, observation $\mathbf{o}_{t}$, and robot state $\mathbf{s}_{t}$, \textit{i.e.}, $\mathbf{a}_{t} = \pi_{\theta}(l, \mathbf{o}_{t}, \mathbf{s}_{t})$.
\name{} adopts the mixture-of-transformers architecture~\cite{liang2024mixture}.
It processes the observation images from multiple camera views and the language instruction with a pre-trained vision-language model (VLM), \textit{i.e.}, Qwen2.5-VL-3B-Instruct~\cite{bai2025qwen2}, and predicts the action chunk with an action diffusion transformer (DiT)~\cite{peebles2023scalable}.

Specifically, \name{} employs flow matching for action prediction~\cite{black2410pi0, liu2022rectified, lipman2022flow}.
The flow prediction is conditioned on the current robot state $\mathbf{s}_{t}$ and the KV cache outputted from the VLM backbone.
The $k$-length action chunk $\ba_t$ is represented as $k$ tokens and concatenated with the robot state token to create the input token sequence for the action DiT.
The flow matching timestep is injected via the adaptive layer norm (AdaLN)~\cite{perez2018film}.
We apply the causal attention mask in the action DiT to model the temporal dependency inside the action chunk.
To ensure fast inference, the action DiT contains half the number of layers compared to the VLM backbone and utilizes only the KV cache from the latter half of the VLM layers.
In total, \name{} contains 4B parameters.

In our early explorations, we observed frequent instability during training.
Inspired by QK norm~\cite{henry2020query}, we apply additional RMSNorm~\cite{zhang2019root} after the linear layers inside both attention and feed-forward networks (FFN) within the DiT blocks.
This design choice drastically improves the stability across the whole training process.
In addition, we found that it significantly improves the language following capability in our downstream experiments, as shown in Sec.~\ref{sec:experiments}.
\section{Training Recipe}
\label{sec:data_training}

We train the \name{} model on a mixture of data sources, including robot trajectory data for imitation learning, web-scale vision-language data for co-training, and human trajectory data for few-shot generalization.
This training recipe allows \name{} to 1) generalize to novel objects, environments, and instructions, 2) efficiently adapt to unseen settings at a low cost, and 3) perform long-horizon and dexterous tasks robustly.

\subsection{Imitation Learning with Robot Trajectory Data}
We train \name{} with an imitation learning objective by maximizing the log-likelihood of the policy on a set of expert demonstrations $\mathcal{D}$:
\begin{equation}
    \max_\theta \mathbb{E}_{\{\ba_t, \bo_t, \bs_t, l\}\sim \mathcal{D}}\left[ \log \pi_\theta (\ba_t\mid \bo_t, \bs_t, l) \right].
\end{equation}
Specifically, we supervise the action prediction with a flow-matching loss during training:
\begin{equation}
    L_\mathrm{action}(\theta) = \mathbb{E}_{\{\ba_t, \bo_t, \bs_t, l\}\sim \mathcal{D}}\left[ \parallel \mathbf{v}_{\theta}(l, \mathbf{o}_{t}, \mathbf{s}_{t}, \mathbf{a}_{t}^{\tau}) - \mathbf{u}(\mathbf{a}_{t}^{\tau}|\mathbf{a}_{t}) \parallel ^{2}\right].
\end{equation}
where $\tau \sim \mathcal{U}\left(0, 1\right)$ is the flow matching timestep and $t$ denotes the episode timestep.
$\mathbf{a}_{t}^{\tau} = (1 - \tau) \epsilon + \tau \mathbf{a}_{t}$ is the noisy action chunk where $\epsilon \sim \mathcal{N}(\mathbf{0}, \mathbf{I})$ is a random noise, and $\mathbf{u}(\mathbf{a}_{t}^{\tau}|\mathbf{a}_{t}) = \mathbf{a}_{t} - \epsilon$ is the ground-truth label for flow prediction.
To accelerate training, we compute the flow-matching loss on multiple sampled flow matching timesteps for one forward pass of the VLM backbone~\cite{li2024autoregressive}.
During inference, the action chunk is initialized from a random noise $\mathbf{a}_{t}^{\tau=0} \sim \mathcal{N}(\mathbf{0}, \mathbf{I})$ and integrated from $\tau=0$ to $\tau=1$ with the Euler method, \textit{i.e.}, $\mathbf{a}_{t}^{\tau + \Delta \tau} = \mathbf{a}_{t}^{\tau} + \mathbf{v}_{\theta}(l, \mathbf{o}_{t}, \mathbf{s}_{t}, \mathbf{a}_{t}^{\tau})\Delta \tau$.
We set $\Delta \tau = 0.2$ in the experiment.

\begin{figure}[!t]
    \centering
    \includegraphics[width=0.98\linewidth]{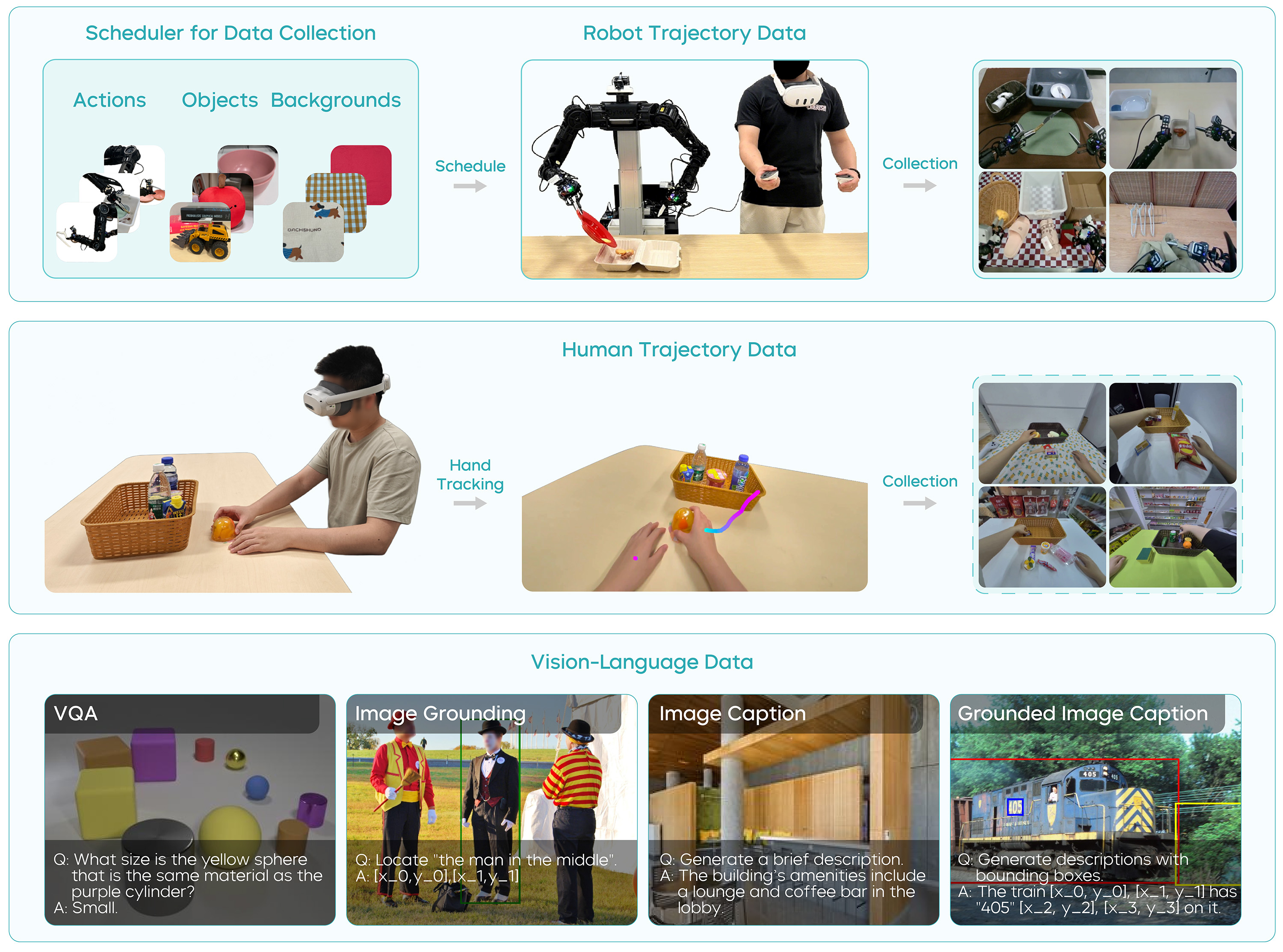}
    \caption{
        \textbf{The \name{} Data.}
        We leverage three types of data during training: robot trajectory data (top), human trajectory data (middle), and vision-language data (bottom).
    }
    \label{fig:data}
\end{figure}

We collect real robot trajectories with teleoperation.
To make the collection process more controllable and maximize the data diversity, we develop a data collection scheduler (Fig.~\ref{fig:data}) to inform teleoperators with 1) the action to perform, 2) the object combination, and 3) the background setting.
At the beginning of each trajectory collection, the system generates a new configuration for the teleoperator to arrange the environment accordingly.
The implementation of the scheduler enables us to effectively manage the overall data distribution and thoroughly randomize the collected data, greatly enhancing the richness and variability of the dataset.
In addition, post-collection quality checks are conducted to refine the dataset by filtering out invalid and low-quality data.

Previous work~\cite{kim2025fine} indicates that policies can take advantage of spurious correlations from multiple viewpoints in action prediction instead of properly attending to the language condition.
To address this issue, we incorporate a ``task status'' as an additional action dimension for auxiliary supervision.
The task status can be one of the following: \texttt{Ongoing} (0), \texttt{Terminated} (1), \texttt{Invalid} (-1).
The \texttt{Ongoing} status indicates that the robot is in progress with the task, while the \texttt{Terminated} status signifies that the robot has successfully completed the task.
The \texttt{Invalid} status indicates that the given instruction is invalid given the current observation.
For example, if there are no knives on the table, ``\textit{put the knife into the woven basket}'' is considered invalid.
During training, we randomly replace the language instruction with an invalid instruction, and train the model to predict the \texttt{Invalid} status without supervision on the other dimensions of the action chunk.
This design forces the action DiT to attend to the language instruction and estimate the task status, substantially improving the language following capabilities, as shown in Sec.~\ref{sec:exp:table}.

\subsection{Co-Training with Vision-Language Data}
To endow \name{} with the generalization capabilities to follow out-of-distribution (OOD) instructions, we jointly train \name{} on both robot trajectories and vision-language data (Fig.~\ref{fig:arch})~\cite{brohan2023rt}.
The robot trajectory data trains both the VLM backbone and the action DiT with the flow-matching objective.
The vision-language data trains only the VLM backbone with the next-token-prediction objective.
For simplicity, we dynamically mix vision-language data with robot trajectories across mini-batches with equal weights.
As a result, the co-training objective is the sum of the next-token-prediction loss and the flow-matching loss.

Through co-training with vision-language data, \name{} is able to effectively generalize to unseen objects and understand novel semantics of complex concept in a zero-shot manner.
We curate a large vision-language dataset from a mixture of data sources~\cite{guo2025seed1, li2024llava, johnson2017clevr, rasheed2024glamm, xue2024xgen}.
This curated dataset covers a wide range of tasks (Fig.~\ref{fig:data}), including image captioning, visual question answering, image grounding, and interleaved grounded image captioning.
We also develop a filtering and re-annotation pipeline to improve the quality of the dataset for effective co-training.
The co-training not only helps \name{} maintain the strong vision-language capabilities from the pre-trained VLM, but also enables the action DiT to leverage these capabilities in action prediction, effectively improving the generalization capabilities in downstream manipulation tasks.

\subsection{Few-Shot Generalization with Human Trajectory Data}
\name{} is a versatile VLA model that can be easily fine-tuned to adapt to novel settings.
However, collecting real-robot trajectories is both time-consuming and costly.
Recent advances in VR devices and hand tracking technology create a promising opportunity to learn actions directly from human trajectory data~\cite{kareer2024egomimic, qiu2025humanoid, hoque2025egodex}.
In this report, we extend the efficient fine-tuning capabilities of \name{} to the challenging setting of few-shot learning from minimal human trajectories.
Specifically, given a novel setting, we collect a small amount of human trajectory data with PICO 4 Ultra Enterprise.
Human trajectories can be efficiently collected with VR devices at a rate of approximately 450 trajectories per hour, substantially outpacing the teleoperated robot trajectory collection, which collects about 250 trajectories per hour.
This efficiency facilitates rapid and cost-effective adaptation to novel settings.

Concretely, the collected human trajectory data contains egocentric videos and human hand trajectories.
We use the same labeling pipeline for robot trajectories to label human trajectory data with languages.
After completing the first stage of training with vision-language data and robot trajectories, we incorporate the human trajectory data and perform co-training on all three types of data.
Unlike robot trajectories, human trajectory data contains only an egocentric view and hand trajectories without arm joint states or gripper states.
We consequently pad blank images for the missing wrist views and train the model with only the hand trajectory for human trajectory data.
\section{Hardware \& System}
\label{sec:hardware_systems}

\subsection{The ByteMini Robot}
\begin{figure}
    \centering
    \includegraphics[width=\linewidth]{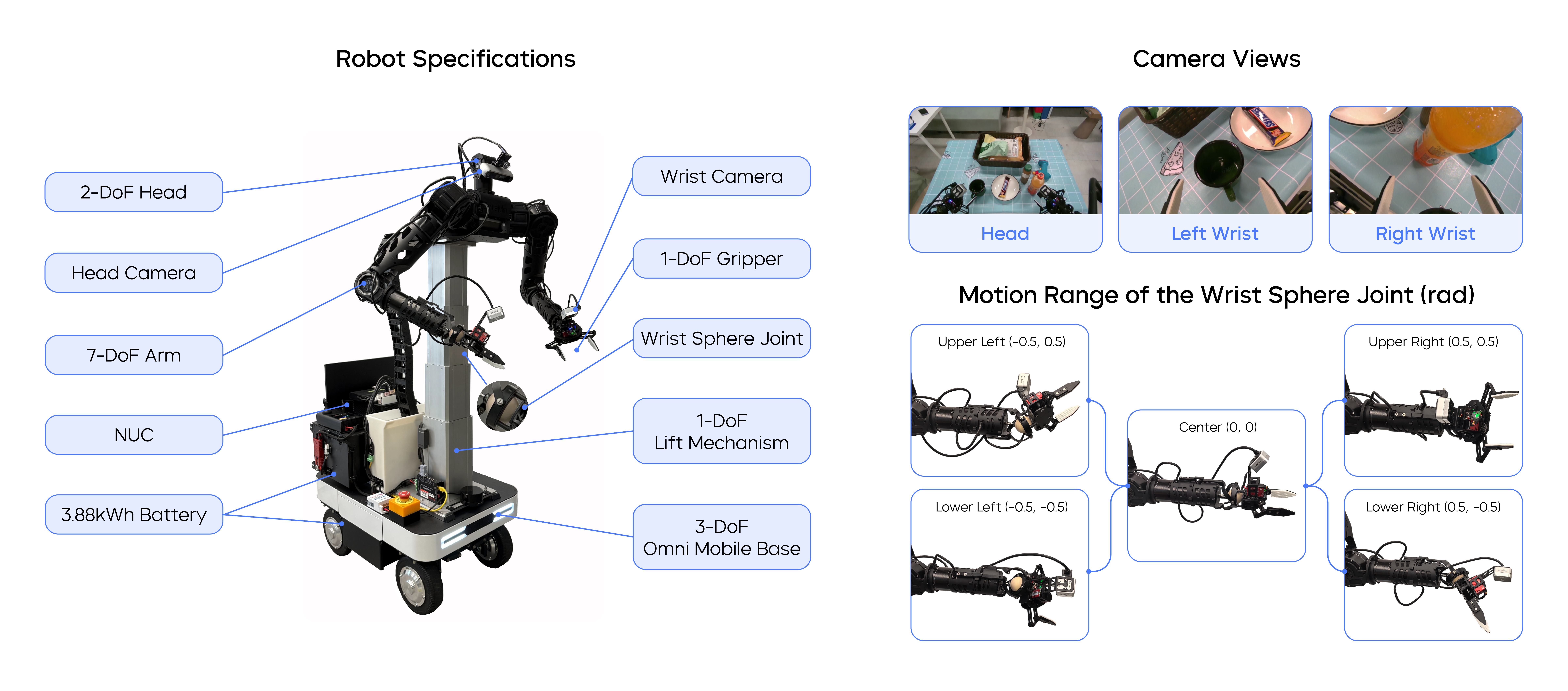}
    \caption{\textbf{The ByteMini Robot.} We show the robot specifications, multi-camera views, and motion range of the unique wrist sphere joint.}
    \label{fig:bytemini_robot}
\end{figure}

The ByteMini robot (Fig.~\ref{fig:bytemini_robot}) is deployed for data collection and policy rollout.
This 22-DoF bi-manual mobile robot is specifically designed with three core objectives: flexible manipulation, high reliability, and user-friendliness.

\paragraph{Flexible manipulation.}
The 7-DoF unbiased robotic arm, featuring a unique sphere wrist joint configuration~\cite{gosselin1993kinematic}, achieves human-like dexterity.
The compact sphere wrist design (Fig.~\ref{fig:bytemini_robot}) overcomes the critical limitation of traditional SRS-configured arms~\cite{nenchev2004singularity}, whose non-compact wrist dimensions hinder flexible operation in confined spaces.
The arm elbow is specifically engineered to allow large internal adduction to 2.53 rad, enabling the two arms to execute delicate operations within the area of the robot chest.

\paragraph{High reliability.}
The demanding workload during data collection and policy rollout requires ByteMini to possess extremely high stability and consistency.
We leverage an omni mobile platform integrated with a lift mechanism to achieve spatial mobility and vertical height adjustment stably.
To further enhance reliability and ensure motion consistency, the actuators in the arms are designed based on the Quasi Direct Drive (QDD) principle~\cite{katz2019mini}, which is well-known for its stability and high transparency.

\paragraph{User-friendliness.}
To improve usability, we integrate a portable screen and an NUC on the robot, supported by a dual-lithium battery that delivers extended endurance of over 10 hours across diverse scenarios.
Additionally, ByteMini is equipped with a wireless E-stop, enabling rapid response to critical situations.

We mount RGBD cameras on the head and the two wrists. 
The wrist cameras enable close-up observation for fine manipulation.

\subsection{System \& Control}
\paragraph{Whole-body compliance control.}
A whole-body compliance control framework~\cite{sentis2006whole} treats all degrees of freedom (DoFs) as a holistic structure to retarget arbitrary teleoperated human motion to feasible robot motion.
Manipulability optimization, singularity avoidance, and physical joint limits are simultaneously addressed within a real-time optimal control problem to maximize robot dexterity.
This generates fluid and continuous motion across large workspaces for diverse long-horizon manipulation tasks, producing high-quality expert trajectories for policy training.
The compliant force controller enables highly dynamic motion and physical interaction with the environment, enhancing safety and data collection efficiency.

\paragraph{Whole-body teleoperation.}
During teleoperated data collection, whole-body retargetting via Meta VR Quest provides intuitive and user-friendly control to directly map human motions to robot end-effectors.
Teleoperators can simultaneously control arm, lift mechanism, gripper, and mobile base movements, facilitating seamless data collection for complex long-horizon tasks in the real world.

\textbf{Trajectory optimization for policy rollout.}
We use the predicted action chunk to control 19 DoFs of the robot (excluding the 3 DoFs from the lift mechanism and head) for policy rollout.
We incorporate pure pursuit~\cite{pursuit} and trajectory optimization to enhance the stability and smoothness of the trajectories generated by \name{} during policy rollout.
The real-time parameterized optimization minimizes jerk and ensures seamless transition between waypoints and across trajectories.

\section{Experiments}
\label{sec:experiments}

We conduct extensive real-world experiments to comprehensively evaluate the performance of \name{}.
Throughout experiments, we aim to answer four questions:
\begin{itemize}[left=0pt]

    \item Does \name{} strictly follow instructions, including those that are unseen during training?
    
    \item Can \name{} generalize to out-of-distribution settings, including novel objects, environments, and instructions?
    
    \item Is \name{} capable of performing few-shot learning from human trajectory data and transfer to the robot embodiment?
    
    \item Can \name{} effectively learn a robust policy capable of performing long-horizon and dexterous task?
\end{itemize}
We consider three tasks: generalizable pick-and-place, long-horizon table bussing, and dexterous cloth hanging.
Please visit the \href{https://seed.bytedance.com/GR3}{project page} for video demos.
We compare our method with the state-of-the-art method $\pi_0$~\cite{black2410pi0}.
We follow the instruction in the official GitHub repository\footnote{https://github.com/Physical-Intelligence/openpi} and fine-tune $\pi_0$ from the provided base model, which is pre-trained on large-scale robot data, for each of the three tasks individually.

\subsection{Generalizable Pick-and-Place}
\label{sec:exp:ppa}

\begin{figure}
    \centering
    \includegraphics[width=\linewidth]{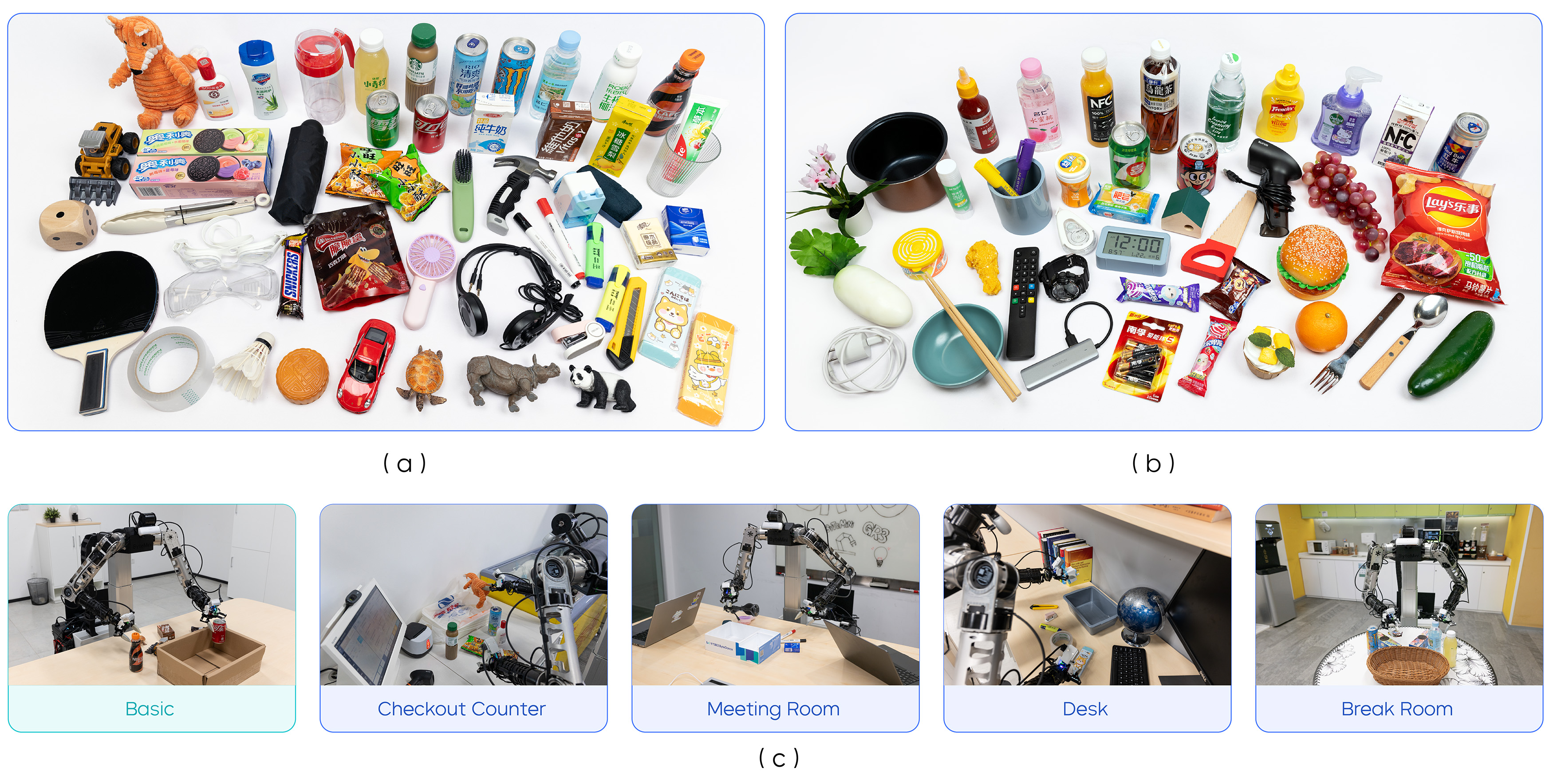}
    \caption{
        \textbf{Experiment Settings of Generalizable Pick-and-Place.}
        (a) Test objects that are seen during training.
        (b) Test objects that are unseen during training.
        (c) The Basic environment is seen during training. The others are out-of-distribution environments that are unseen during training.
    }
    \label{fig:ppa-setting}
\end{figure}

To assess the generalization capabilities of \name{} in out-of-distribution settings, we evaluate on a generalization-focused pick-and-place task.
In total, we collected 35k robot trajectories covering 101 objects, accounting for a total of 69 hours for this task.
We annotate the robot trajectories with the ``\textit{put A into B}'' instruction, where A is the object category and B is the container.
For the baseline model, we fine-tune $\pi_0$ with these robot trajectory data.
For \name{}, we co-train the model with both the robot trajectory data and vision-language data.
During training, we augment the images of the robot trajectories with photometric augmentations to improve the robustness against the changing environment.
We also compare with a variant of our method, \name{} w/o Co-Training, which trains the model with \textit{only} robot trajectories.
This ablation study helps us evaluate the impact of vision-language co-training and identify the specific benefits it contributes to the model performance.

\paragraph{Settings.}
We evaluate under four different settings: 1) Basic, 2) Unseen Environments, 3) Unseen Instructions, and 4) Unseen Objects.
In Basic, we evaluate in an environment that is seen during training.
We evaluate with 54 objects that are seen during training (Fig.~\ref{fig:ppa-setting}(a)) to test the basic instruction following capabilities.
In Unseen Environments, we evaluate with the same set of objects in four different real world environments that are unseen during training: a checkout counter, a meeting room, a desk, and a break room (Fig.~\ref{fig:ppa-setting}(c)).
The layouts of objects are kept consistent with the Basic setting.
In Unseen Instructions, we prompt the model with instructions requiring complex concept understanding, \textit{e.g.}, ``\textit{put the left coke into the carton}'' and ``\textit{put the animal with tentacles into the carton}''.
In Unseen Objects, we evaluate using 45 objects that are unseen in the robot trajectory data (Fig.~\ref{fig:ppa-setting}(b)).

We assess the model performance with the instruction following (IF) rate and the success rate, which measure the ability of the model to follow instructions and its overall performance on task completion, respectively.
For the IF rate, a trial is considered successful if the robot correctly approaches the object specified by the given instruction.
For the success rate, a trial is considered successful if the robot puts the target object into the container.
For both metrics, a higher score reflects a stronger capability.

\begin{figure}
    \centering
    \includegraphics[width=\linewidth]{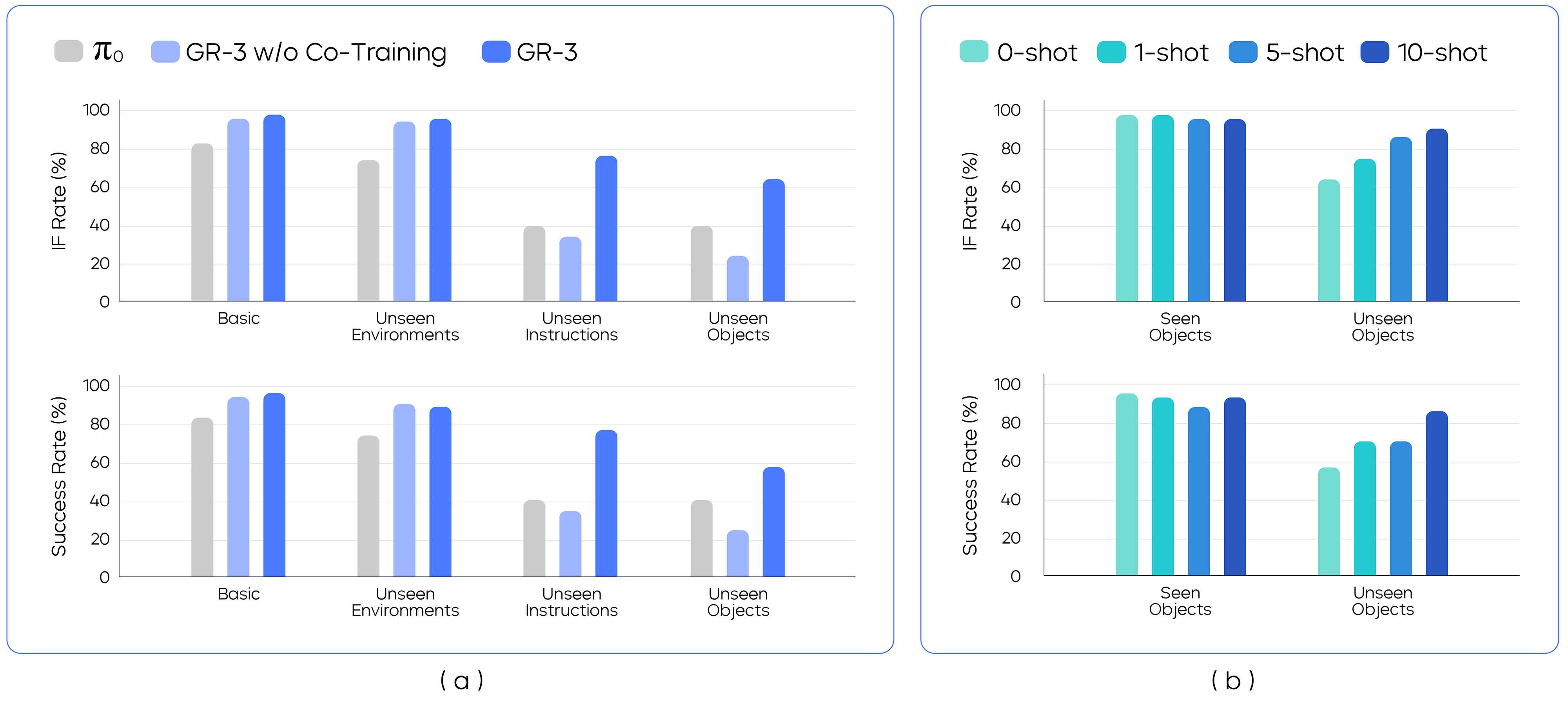}
    \caption{
        \textbf{Experiment Results of Generalizable Pick-and-Place.}
        (a) Results on generalizable pick-and-place under four different settings.
        (b) Results on few-shot generalization with human trajectories.
    }
    \label{fig:ppa-result}
\end{figure}

\paragraph{Basic Instruction Following.}
In Basic and Unseen Environments, we split the 54 seen objects into nine mini-batches with six objects per batch.
In each rollout, we prompt the model to pick one object from \textit{all} the six objects according to the given instruction.
To guarantee the comparability of results across different models, we position the objects according to a pre-captured mask, ensuring that the object layout of a mini-batch remains as consistent as possible during evaluation.

As shown in Fig. \ref{fig:ppa-result}(a), \name{} surpasses $\pi_{0}$ in both Basic and Unseen Environments in terms of the IF rate and success rate.
The modest performance degradation between Basic and Unseen Environments highlights the robustness of \name{} against environment changes.
Additionally, we observe no significant performance differences between \name{} and \name{} w/o Co-Training in these two settings, suggesting that co-training does not impact the model performance on seen objects.

\paragraph{Generalizable Instruction Following.}
In Unseen Instructions, we aim to test the capabilities to understand abstract concepts relating to sizes, spatial relationships, and common-sense knowledge.
Example instructions include ``\textit{put the coke next to the sprite into the carton}'', ``\textit{put the largest object into the carton}'', and ``\textit{put the marine animal into the carton}''.
These instructions are unseen in the robot trajectory data and require the model to reason the complex semantic in the instructions.
In Unseen Objects, we split the 45 unseen objects into nine mini-batches with five objects per batch, \textit{i.e.}, the model needs to select one object from five objects per rollout.
This setting is particularly challenging as among the 45 objects, more than $70\%$ of them are from categories that are unseen in the robot trajectory data.

As shown in Fig.~\ref{fig:ppa-result}(a), \name{} surpasses $\pi_0$ in both settings by a large margin, highlighting its superior generalization capabilities.
It boosts the success rate from 40\% to 77.1\% in Unseen Instructions and from 40\% to 57.8\% in Unseen Objects.
\name{} also substantially outperforms \name{} w/o Co-Training in both settings, indicating that co-training with vision-language (VL) data contributes to the strong generalization capabilities.
The VLA model effectively transfers rich knowledge from large-scale VL data into policy learning and enables powerful zero-shot capabilities in novel settings.
We also observe that training \name{} with only robot trajectories yields an inferior performance compared to the $\pi_0$ baseline.
We hypothesize that the performance superiority of $\pi_0$ stems from its large-scale cross-embodiment pre-training~\cite{black2410pi0}.

\paragraph{Few-Shot Generalization from Human Trajectory Data.}
We also evaluate the few-shot generalization capabilities~\cite{wang2020generalizing,kang2019few} with human trajectories collected with VR devices.
This is challenging for the reason that 1) the model needs to learn from cross-embodiment data and 2) the data is scarce.
Specifically, we collected up to 10 human trajectories per object for the 45 unseen objects in the setting of Unseen Objects (Fig. \ref{fig:ppa-setting}(b)).
The overall duration of the 450 human trajectories is about 30 minutes.
We incrementally train \name{} based on the checkpoint trained with robot trajectories and VL data.
We perform co-training for an additional 20k steps by further incorporating human trajectory data alongside the robot trajectories and VL data.

We evaluate the performance under different few-shot settings (1-, 5-, and 10-shot) on both seen and unseen objects (Fig.~\ref{fig:ppa-result}(b)).
Comparing to the zero-shot performance of the base model, we are able to continue improving the IF rate and success rate with more human trajectory data on unseen objects and boost the success rate from 57.8\% to 86.7\% with only 10 human trajectories per object.
Additionally, we notice that there is no evident performance drop on seen objects, indicating a promising sample-efficient and cost-effective fine-tuning strategy to adapt pre-trained VLA models to downstream novel settings.

\subsection{Long-Horizon Table Bussing}
\label{sec:exp:table}

\begin{figure}[!t]
    \centering
    \includegraphics[width=\linewidth]{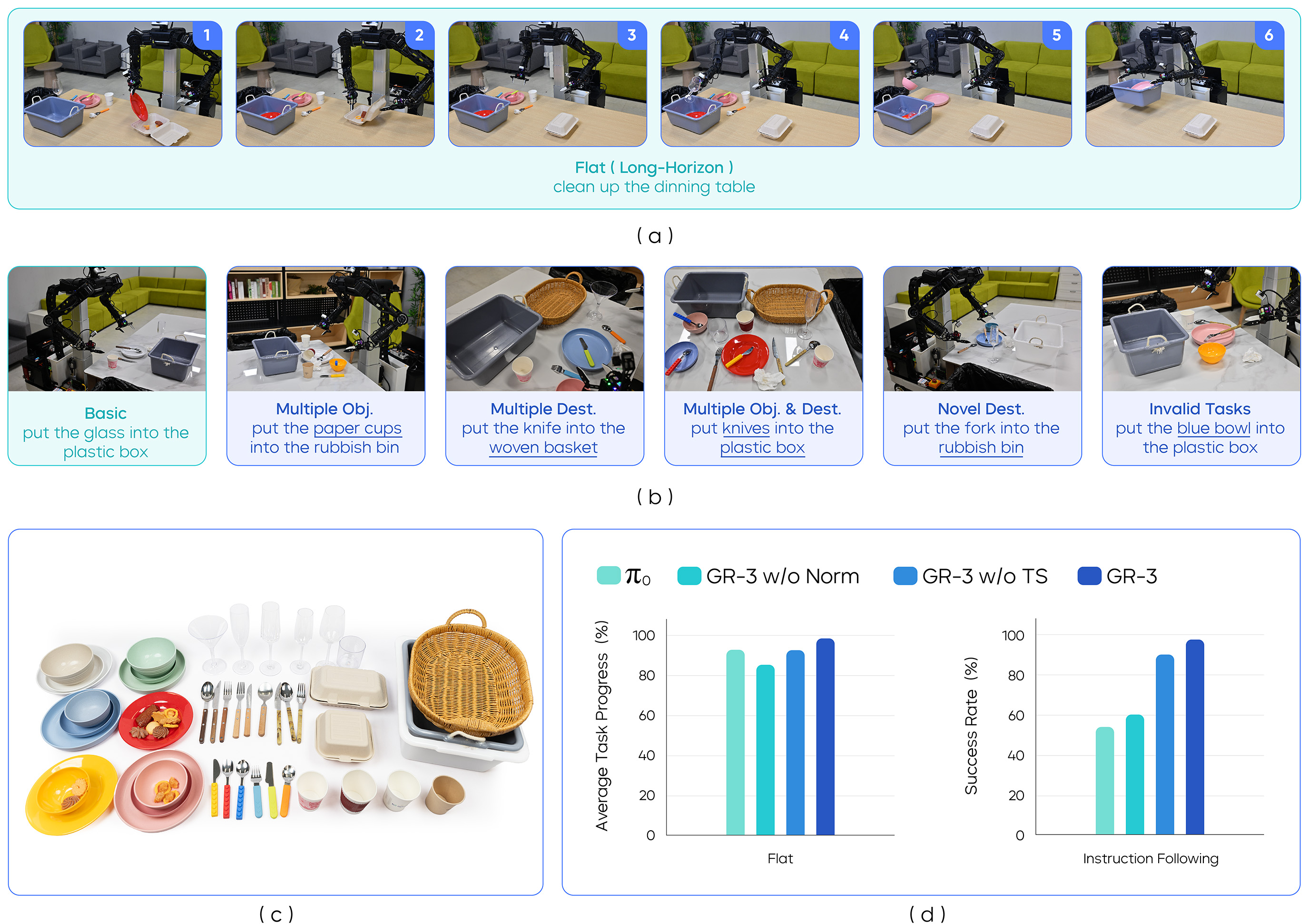}
    \caption{
        \textbf{Experiment Settings \& Results of Table Bussing.} 
        (a) Flat:
        the robot is required to perform long-horizon table bussing in a single run.
        (b) Instruction following (IF): 
        the robot is prompted with multiple sub-task descriptions in a roll. 
        (c) Test objects.
        (d) Results on the flat and instruction following (IF) settings. 
    }
    \label{fig:table_bussing}
\end{figure}

We perform experiments on a table bussing task to evaluate the robustness of \name{} on handling long-horizon manipulation (Fig.~\ref{fig:table_bussing}).
In this task, the robot is required to clean up a table with messy utensils, food, to-go box, and a plastic bussing box.
To complete the task, the robot needs to 1) pack the food into the to-go box, 2) put all the utensils into the bussing box, and 3) put all the trash into the rubbish bin.
Given the large workspace, the robot needs to move its mobile base from the to-go box to the bussing box to accomplish the whole task (Fig.~\ref{fig:table_bussing}(a)).
We evaluate the models in a flat setting and an instruction-following (IF) setting.

\paragraph{The Flat Setting.}
In this setting, we prompt the robot with a general task instruction, ``\textit{clean up the dining table}'', to complete the entire task autonomously within a single run (Fig.~\ref{fig:table_bussing}(a)).
The flat setting helps us evaluate the robustness of the model on handling long-horizon tasks.
We use the average task progress, which calculates the ratio of successfully completed sub-tasks to the total number of sub-tasks, as the evaluation metric.
A value of 1.0 indicates complete success and a fractional value corresponds to partial success.
In total, we evaluate on five different sets of objects for this setting.

\paragraph{The Instruction-Following (IF) Setting.}
In this setting, we further evaluate how the model follows instructions.
We prompt the robot with multiple sub-task descriptions in a roll, \textit{e.g.}, ``\textit{put the paper cup into the rubbish bin}'', to clean up the table.
The robot performs each sub-task starting from a home position.
We use the average sub-task success rate as the evaluation metric.
In total, the IF setting covers six different sets of instructions (Fig.~\ref{fig:table_bussing}(b)):
\begin{itemize}[left=0pt]

    \item \textbf{Basic}: The object layout closely resembles those in the training data.
    
    \item \textbf{Multiple Objects}: We add multiple instances of a subset of object categories into the scene. We include instructions to command the robot to put all the instances belonging to these categories into the bussing box or rubbish bin.
    
    \item \textbf{Multiple Destinations}: We add a woven basket into the scene and command the robot to put utensils into either the woven basket or the bussing box.
    
    \item \textbf{Multiple Objects \& Destinations}: We combine the above two settings and instruct the robot to move all the instances of an object category to one of the two destinations.
    
    \item \textbf{Novel Destinations}: The robot is required to move objects to a destination that does not appear with the object in the training data, \textit{e.g.}, ``\textit{put the fork into the rubbish bin}''.
    
    \item \textbf{Invalid Tasks}: In real-world application, the robot needs to deal with complex instructions among which some may be invalid. For example, if there are no blue bowls on the table, ``\textit{put the blue bowl into the plastic box}'' would be considered invalid. In such scenario, we want the policy to reject from performing wrong valid tasks~\cite{milli2017should}. In this setting, we prompt the model with tasks that can not be completed with the given observation. The trial is considered successful only if the model refrains from manipulating any objects within 10 seconds.

\end{itemize}

\begin{figure}[!t]
    \centering
    \includegraphics[width=\linewidth]{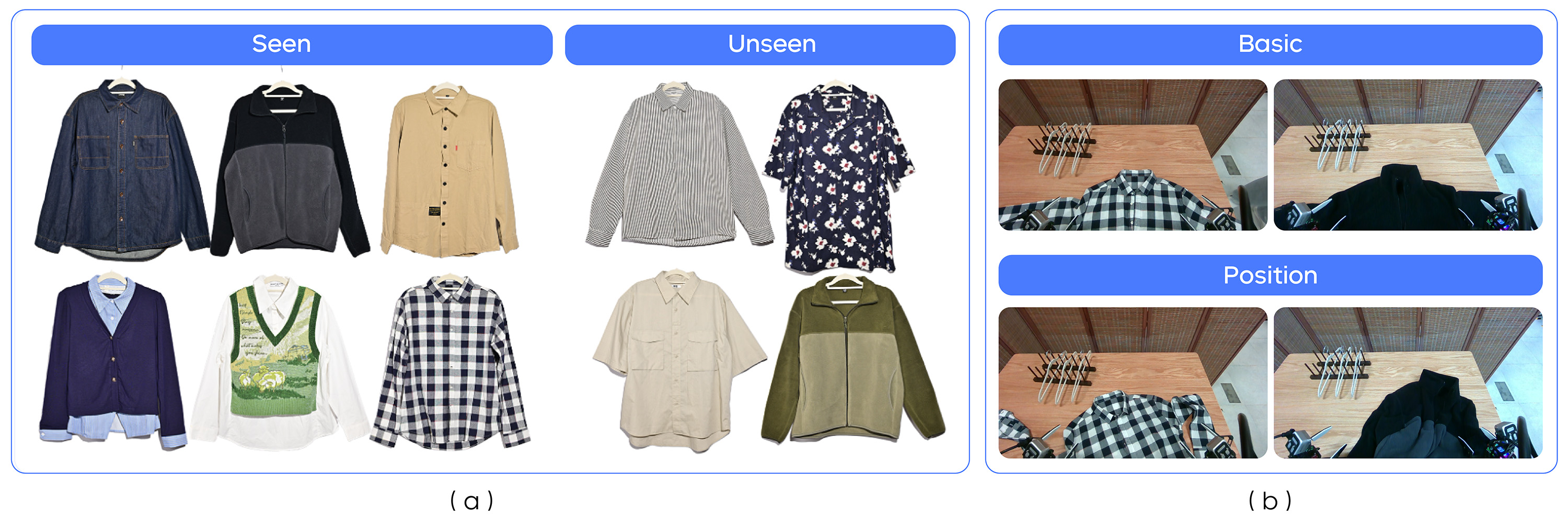}
    \caption{
        \textbf{Experiment Settings of Dexterous Cloth Manipulation.}
        (a) Seen and unseen clothes in the test set.
        (b) The Basic and Position settings.
    }
    \label{fig:cloth_exp_setting}
\end{figure}

\paragraph{Implementations.}
In total, we collected approximately 101 hours of robot trajectories for this task.
For the baseline method, we fine-tune $\pi_0$ on these robot trajectories.
For \name{}, we co-train on both the robot trajectories and vision-language (VL) data.
We also perform ablation studies with two variants of our method, \name{} w/o Norm and \name{} w/o Task Status (TS).
\name{} w/o Norm removes the introduced RMSNorm in the attention and FFN of the DiT blocks.
\name{} w/o TS does not incorporate task status during training.
For all methods, we train two separate models, a flat version and an IF version, for the two settings, respectively.
For the flat version, we randomly sample between the general task and sub-tasks as the language instruction.
For the IF version, we use only sub-tasks as the instruction during training.

\paragraph{Results.}
As shown in Fig.~\ref{fig:table_bussing}(d), \name{} outperforms $\pi_0$ in both settings, especially in the IF setting (53.8\% vs 97.5\% in terms of success rate).
While $\pi_0$ is able to perform long-horizon table bussing, it struggles with instruction following, especially in out-of-distribution scenarios.
It is not able to distinguish between forks and spoons.
And in Novel Destinations, it puts objects into containers that appear together with the objects in the training data instead of following the given instructions.
On the other hand, \name{} strictly follows instructions in all six test sets.
It generalizes well to multiple objects and destinations and is able to refain from performing wrong tasks in Invalid Tasks.

Removing the RMSNorm hurts the performance in both settings, especially in the IF setting.
\name{} w/o Norm is not able to follow instructions well.
In particular, it is not able to generalize to novel destinations.
These results underscore the crucial role of RMSNorm in enhancing the instruction following capabilities.
Without the task status, the IF capabilities also drop, highlighting the effectiveness of task status in aiding the VLA model to follow instructions.

\subsection{Dexterous Cloth Manipulation}
\label{sec:exp:cloth}
In this experiment, we evaluate \name{} on dexterous manipulation of deformable objects.
In particular, we challenge the model to hang clothes onto a drying rack with cloth hangers (Fig.~\ref{fig:capability}).
In this task, the robot needs to 1) pick up the hanger, 2) place the clothes onto the hanger, and 3) hang the clothes on the drying rack.
In the final step, the robot needs to rotate its mobile base from the table to the drying rack to hang the clothes.
In total, we collected 116 hours of robot trajectories for this task.
We train $\pi_0$ on these data.
For \name{}, we co-train on these robot trajectories and vision-language data.
We perform evaluation under three different settings: Basic, Position, and Unseen Instances.

\begin{figure}[!t]
    \centering
    \includegraphics[width=\linewidth]{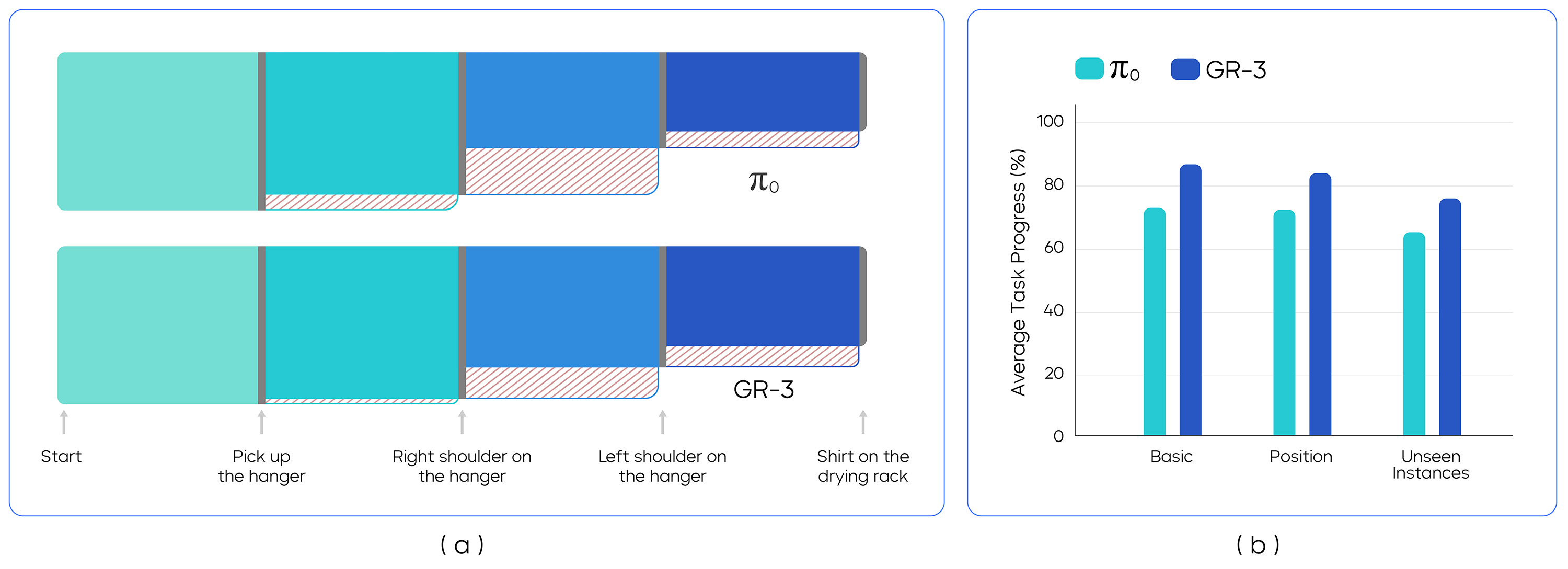}
    \caption{
        \textbf{Experiment Results on Dexterous Cloth Manipulation.}
        (a) Sankey diagram of success (solid) and failure (hatch) across entire rollouts of the Basic setting.
        (b) Average task progresses of $\pi_0$ and \name{} in the three evaluation settings.
    }
    \label{fig:cloth_exp_result}
\end{figure}

\paragraph{Settings.}
For Basic, we evaluate on six clothes that are seen during training.
The clothes are positioned similarly as in the training data.
For Position, we rotate and crumple the clothes as shown in Fig.~\ref{fig:cloth_exp_setting}(b).
The Position setting evaluates the model's robustness in handling challenging clothes layouts.
In Unseen Instances, we evaluate the model's ability to generalize to clothes that are not seen during training.
Specifically, we evaluate with four unseen clothes (Fig.~\ref{fig:cloth_exp_setting}(a)).
While all the clothes are long-sleeved in the training data, two of the unseen clothes in the test set are short-sleeved.
We use average task progress as the evaluation metric, where a full success -- hanging the shirt on the drying rack -- corresponds to 1.0.
The entire process is divided into four key milestones: 1) picking up the hanger, 2) placing the right shoulder on the hanger, 3) placing the left shoulder on the hanger, and 4) hanging the shirt on the drying rack (Fig.~\ref{fig:cloth_exp_result}(a)).
Each milestone contributes a fractional score towards the overall task progress.

\paragraph{Results.}
Results are shown in Fig.~\ref{fig:cloth_exp_result}.
\name{} outperforms $\pi_0$ in all three evaluation settings.
It achieves an average task progress of 86.7\% and 83.9\% in Basic and Position, respectively, showcasing its proficiency in handling complex dexterous tasks and its robustness to positional variations.
Furthermore, \name{} is able to generalize to unseen clothes with novel patterns and sleeve length, achieving an average task progress of 75.8\%.
To probe into the rollout process, we show the Sankey diagram of success and failure across the four milestones of the Basic setting in Fig.~\ref{fig:cloth_exp_result}(a).
The most challenging part for both model is to place the left shoulder on the hanger after the right shoulder.
This is because the robot needs to pull out the left collar, which is often folded behind the hanger, for grasping while holding the hanger.
Another failure mode occurs when the hanger slips away from the gripper during the course of placing the left shoulder on it, leading to failures in the last step.
\section{Related Work}

\paragraph{Generalist Manipulation Policies.}
Building generalist manipulation policies capable of following instructions to effectively interact with the physical world has been a long-standing challenge in robotics research~\cite{brohan2022rt, brohan2023rt, kimopenvla, black2410pi0, intelligence2025pi_, bjorck2025gr00t, team2024octo, li2023vision, li2024towards, reuss2024multimodal, bharadhwaj2024roboagent, shukor2025smolvla, barreiros2025careful}.
Prior works~\cite{nair2022r3m, radford2021learning, xiao2022masked, karamcheti2023language} propose to learn representations from large-scale data for downstream policy learning, enabling more robust robot behaviors in complex tasks.
Recent advancements in vision-language-action (VLA) models adopt various approaches to enhance policy generalization and improve manipulation capabilities.
A popular line of works~\cite{o2024open, team2024octo, kimopenvla, black2410pi0, pertsch2025fast, intelligence2025pi_, doshi2024scaling, wang2024scaling, liu2024rdt, qu2025spatialvla, li2024cogact} train policies with cross-embodiment data that encompasses trajectories collected from different robot embodiments~\cite{o2024open, khazatsky2024droid, dasari2019robonet, walke2023bridgedata, ebert2021bridge}.
Along with the real-world robot trajectories, others~\cite{brohan2023rt, black2410pi0, intelligence2025pi_} leverage pre-trained vision-language models to develop robot policies, showcasing strong capabilities in generalization to unseen settings.
Another way to improve generalization is to perform future prediction on web-scale video datasets~\cite{wu2023unleashing, cheang2024gr, hu2024video, li2025gr, black2023zero, du2023learning} or learn latent actions~\cite{bjorck2025gr00t, bu2025agibot, ye2024latent} from action-less videos.
In this work, we introduce \name{}, a VLA model that is jointly trained on both robot trajectories and vision-language data, and can be efficiently fine-tuned with few-shot human trajectories.
Throughout extensive experiments, we show that \name{} is able to 1) strictly follow instructions and generalize to novel objects, environments, and instructions, 2) efficiently adapt to novel settings from few-shot human trajectory data, and 3) perform long-horizon and dexterous tasks with high robustness.

\paragraph{Multi-Modal Co-Training for Robot Manipulation.}
Collecting real-world robot trajectories is costly and time-consuming.
As a result, broadening data sources becomes critical when it comes to scaling up policy training~\cite{kaplan2020scaling, lin2024data}.
A popular approach is to initialize policies from pre-trained vision encoders~\cite{nair2022r3m, xiao2022masked, majumdar2023we, karamcheti2023language, radford2021learning} or more recently, pre-trained vision-language models~\cite{black2410pi0, intelligence2025pi_, brohan2023rt, bjorck2025gr00t}.
Built upon this framework, it is natural to incorporate multi-modal data during training besides robot trajectories~\cite{reed2022generalist, yang2025magma, brohan2023rt, intelligence2025pi_, pertsch2025fast}.
In particular, GATO~\cite{reed2022generalist} builds a generalist agent capable of performing a wide range of tasks, including image captioning and block stacking with a real robot.
It converts data from different modalities into a sequence of tokens and trains a large transformer for next-token prediction.
RT-2~\cite{brohan2023rt} demonstrates that co-fine-tuning large vision-language models on both robot trajectories and vision-language data significantly boosts generalization.
By co-training on heterogeneous data, $\pi_{0.5}$~\cite{intelligence2025pi_} showcases exceptional generalization across environments and objects, enabling effective deployment in real-world unseen scenarios.
In this work, \name{} adopts a similar co-training strategy.
We curate a well-designed web-scale vision-language dataset from multiple data sources and perform large-scale co-training on a wide range of vision-language tasks.
Through large-scale co-training, \name{} showcases strong capabilities in zero-shot generalization to unseen objects and complex instructions that require abstract concept understanding.

\paragraph{Leveraging Human Data for Policy Training.}
To improve data efficiency, incorporating human data into policy training has become a popular approach in robotics research.
A popular line of works~\cite{mendonca2023structured, bharadhwaj2024track2act, bahl2023affordances, wen2023any} extract different types of representations from human videos to enhance policy training.
\citet{wu2023unleashing} and \citet{cheang2024gr} propose to leverage large-scale human videos~\cite{grauman2022ego4d, damen2018scaling, goyal2017something} for generative video pre-training.
Recently, with the development of hand tracking and VR devices, \citet{qiu2025humanoid} and \citet{kareer2024egomimic} show that human videos with hand trajectories can improve policy performance on robot embodiments via co-training with a small amount of robot data.
In this report, we follow this line of research and show that \name{} can effectively adapt to novel settings through few-shot learning from human trajectories.

\section{Limitations \& Conclusions}
\label{sec:conclusions}
\paragraph{Limitations and future work.} Despite the powerful performance in challenging tasks, \name{} has limitations.
While it showcases strong generalization capabilities, it makes mistakes in following unseen instructions involving novel concepts and objects, and struggles with grasping objects with unseen shapes.
We plan to scale up the model and training data to continue improving the model performance on handling novel scenarios.
In addition, similar to all imitation learning methods, \name{} can be stuck in out-of-distribution states during rollouts and fails to recover from failure.
In the future, we plan to incorporate reinforcement learning (RL) to strengthen robustness on complex and dexterous tasks and optimize performance beyond the limitation of imitation learning.

\paragraph{Conclusions.} In this report, we introduce \name{}, a powerful vision-language-action (VLA) model that outputs actions to control a bi-manual mobile robot.
We carefully study the model architecture and develop a comprehensive training recipe that combines co-training with large-scale vision-language data, efficient few-shot learning from human trajectories, and effective imitation learning from robot trajectories.
Extensive real-world experiments on three challenging tasks showcase that \name{} excels in understanding complex instructions with abstract concepts, generalizes effectively to novel objects and environments, learns efficiently from minimal human trajectories, and performs long-horizon and dexterous tasks with exceptional robustness and reliability.
We hope \name{} can serve as a step towards creating generalist robots capable of assisting humans with diverse tasks in the real world.

\clearpage
\section{Contributions and Acknowledgements}
\label{sec:contributions}

Authors contributed in the following areas are listed by alphabetical order.

\begin{itemize}[left=0pt]
    \item \textbf{Data:} Chilam Cheang, Sijin Chen, Yifeng Li, Yuxiao Liu, Haixin Shi, Yuyang Xiao
    \item \textbf{Model Architecture:} Hongtao Wu, Xin Xiao
    \item \textbf{Training:} Sijin Chen, Xiao Ma, Hongtao Wu, Xin Xiao, Jiafeng Xu, Yichu Yang
    \item \textbf{Evaluation:} Chilam Cheang, Sijin Chen, Yingdong Hu, Wenxuan Ou, Haixin Shi, Hongtao Wu, Xin Xiao, Yichu Yang
    \item \textbf{Robot Development:} Zhongren Cui, Liqun Huang, Hao Niu, Zeyu Ren, Jiawen Tian, Jiafeng Xu
    \item \textbf{Robot System:} Yuxiao Liu, Wenxuan Ou, Wanli Peng
    \item \textbf{Writing:} Sijin Chen, Yuxiao Liu, Xiao Ma, Zeyu Ren, Hongtao Wu, Yichu Yang
    \item \textbf{Team Lead:} Tao Kong, Hang Li

\end{itemize}
We thank Jinming Guo, Xueyang Hu, Zetian Li, Xuguang Wei, Zhaohao Xiao, Degong Yang, Yifan Yang, Zhouqike Yang, and Yifei Zhou for their help on data curation.
We thank Zeyu Long and Tingshuai Yan for their help on model deployment.
We thank Yang Liu and Ming Zhao for their help on robot development and maintenance.
We are sincerely grateful to all the teleoperators and annotators for their dedicated efforts on data collection and annotation.

\clearpage

\bibliographystyle{plainnat}
\bibliography{main}

\end{document}